\documentclass[runningheads]{llncs}
\usepackage[T1]{fontenc}
\usepackage{booktabs}
\usepackage{graphicx}
\usepackage{subcaption}
\usepackage{standalone}

\usepackage{tikz}
\usepackage{pgfplots}
\pgfplotsset{compat=1.18}
\usepackage{amsmath}
\usepackage{amssymb}
\usepackage{listofitems}
\usepackage[outline]{contour}

\usetikzlibrary{arrows.meta, fillbetween, pgfplots.fillbetween}

\contourlength{1.4pt}

\usepackage{xcolor}
\colorlet{myred}{red!80!black}
\colorlet{myblue}{blue!80!black}
\colorlet{mygreen}{green!60!black}
\colorlet{myorange}{orange!70!red!60!black}
\colorlet{mydarkred}{red!30!black}
\colorlet{mydarkblue}{blue!40!black}
\colorlet{mydarkgreen}{green!30!black}
\colorlet{boxbg}{gray!10!white}

\tikzset{
  >=latex, 
  node/.style={thick,circle,draw=myblue,minimum size=22,inner sep=0.5,outer sep=0.6},
  node in/.style={node,green!20!black,draw=mygreen!30!black,fill=mygreen!25},
  node hidden/.style={node,blue!20!black,draw=myblue!30!black,fill=myblue!20},
  node convol/.style={node,orange!20!black,draw=myorange!30!black,fill=myorange!20},
  node out/.style={node,red!20!black,draw=myred!30!black,fill=myred!20},
  connect/.style={thick,mydarkblue},
  connect arrow/.style={-{Latex[length=4,width=3.5]},thick,mydarkblue,shorten <=0.5,shorten >=1},
  node 1/.style={node in}, 
  node 2/.style={node hidden},
  node 3/.style={node out}
}

\usepackage{hyperref}
\usepackage{cleveref}
\newif\ifpreprint
\preprinttrue
\ifpreprint
\renewcommand{\orcidID}[1]{}
\fi
\newcommand{\preprintfootnote}{%
    \ifpreprint
    \begingroup
    \renewcommand{\thefootnote}{}%
    \footnotetext{\footnotesize This preprint has not undergone peer review
    or any post-submission improvements or corrections. The Version of Record
    of this contribution will be published in
    \textit{Computer Aided Systems Theory -- EUROCAST~2026},
        Lecture Notes in Computer Science, Springer.}%
    \endgroup
    \fi
}
\usepackage{xspace}

\newcommand{\eg}{e.g.\xspace}
\usepackage{glossaries}
\usepackage{amsfonts}
\newacronym{mtsad}{MTSAD}{Multivariate Time Series Anomaly Detection}
\newacronym{fm}{FM}{Foundation Model}
\newacronym{if}{IF}{Isolation Forest}

\newacronym{msl}{MSL}{Mars Science Laboratory}
\newacronym{psm}{PSM}{Pooled Server Metrics}
\newacronym{qad}{QAPPD}{Quanser Aero 2 Pick-and-Place Dataset}
\newacronym{smap}{SMAP}{Soil Moisture Active and Passive}
\newacronym{smd}{SMD}{Server Machine Data}
\newacronym{swat}{SWaT}{Secure Water Treatment}
\newacronym{wadi}{WaDi}{Water Distribution}

\usepackage{color}

\begin{document}

    \title{Exploring Zero-Shot Foundation Models for Multivariate%
    Time Series Anomaly Detection}
    \titlerunning{Exploring Zero-Shot FMs for MTSAD}

    \author{
        Martin Uray\inst{1,2}\orcidID{0000-0002-8916-3847} \and
        Saverio Messineo\inst{1}\orcidID{0000-0003-1592-4428} \and
        Roland Kwitt\inst{2}\orcidID{0000-0001-9947-4465} \and
        Stefan Huber\inst{1}\orcidID{0000-0002-8871-5814}
    }
    \authorrunning{M. Uray et al.}
    \institute{
        Josef Ressel Centre for Intelligent and Secure Industrial Automation,\\
        Salzburg University of Applied Sciences, Austria\\
        \email{\{martin.uray, saverio.messineo, stefan.huber\}@fh-salzburg.ac.at}
        \and
        Department of Artificial Intelligence and Human Interfaces,\\
        Paris Lodron University of Salzburg, Austria\\
        \email{roland.kwitt@plus.ac.at}
    }
    \maketitle              
    \preprintfootnote
    \begin{abstract}
        \gls{mtsad} is essential for reliability
        and safety in domains such as industrial process monitoring and
        financial risk management, yet conventional approaches rely on
        application-specific models that are costly to train and hard to scale.
        \glspl{fm}, pre-trained on broad data with strong zero-shot
        generalization, have recently become available for univariate time
        series forecasting, raising the question of whether they can address
        \gls{mtsad} without task-specific training.
        We investigate the zero-shot application of a univariate forecasting
        \gls{fm}, TimesFM, to industrial \gls{mtsad} on the \gls{swat}
        benchmark, evaluating two strategies: treating the \gls{fm} as a
        per-feature forecaster with thresholded prediction errors, and as an
        embedder whose intermediate representations feed standard outlier
        detectors.
        Neither of our proposed setups is competitive with established
        baselines; embeddings reveal only partial separation between normal and
        anomalous segments, insufficient for reliable detection.
        The cause is that the \gls{fm} is too effective at capturing temporal
        dynamics, yielding low error even within fully anomalous windows, so
        persistent anomalies become indistinguishable from normal behavior.
        However, these observations yield valuable insights:
        the error peaks at anomaly boundaries, indicating \glspl{fm} reliably
        detect distribution changes. We conclude that the proposed naive
        zero-shot \glspl{fm} are  unsuitable for \gls{mtsad} but promising for
        change-point detection.

    \end{abstract}
    \glsresetall
    \section{Introduction}\label{sec:introduction}
    \gls{mtsad} plays a vital role in ensuring the reliability and safety of
    numerous application domains, including industrial process monitoring,
    predictive maintenance, fraud detection, and
    cybersecurity~\cite{sarfraz2024}.
    The recorded data in these settings is typically high-dimensional,
    noisy, and
    strongly imbalanced, as anomalous events are rare and heterogeneous.
    Consequently, detection is commonly addressed by models tailored to a
    specific application and dataset — an approach that is costly to develop,
    energy-intensive to retrain, and difficult to scale across environments or
    distribution shifts.

    Forecasting-based methods are among the most widely adopted paradigms for
    \gls{mtsad}~\cite{sarfraz2024}. A model $f_\theta$ is trained to predict
    future observations from a context window, \eg
    $\hat{X}_{t+1} = f_\theta(X_{t-w:t})$, and anomalies are flagged when the
    prediction error $e_{t} = \| \hat{X}_{t+1} - X_{t+1} \|$ exceeds a
    threshold. The model must be trained on in-distribution data, however,
    which perpetuates the per-dataset specialization problem.

    \glspl{fm} — large models pre-trained on broad, diverse data through
    self-supervision — offer a potential escape from this
    cycle~\cite{bommasani2021}.
    They have demonstrated strong zero-shot performance across many domains,
    and dedicated univariate time series \glspl{fm} have recently become
    available~\cite{das2024}. Applying such a model directly to \gls{mtsad}
    would eliminate the need for dataset-specific training while leveraging
    representations learned from large-scale temporal data.
    Despite this appeal, the zero-shot application of \glspl{fm} to industrial
    \gls{mtsad} remains largely unexplored.

    We therefore investigate: \emph{Can a zero-shot univariate time series
    \gls{fm} detect anomalies in industrial multivariate time series without
    any task-specific training?} We make the following contributions:
    \begin{itemize}
        \item We propose and evaluate two complementary zero-shot strategies for
        applying a univariate forecasting \gls{fm} to \gls{mtsad}: using it as
        a per-feature \emph{forecaster} with thresholded prediction errors, and
        as an \emph{embedder} whose intermediate representations feed a standard
        outlier detector.
        \item We show that neither strategy is competitive with established
        baselines on the \gls{swat} benchmark, and identify the root cause: the
        \gls{fm} adapts to anomalous dynamics just as effectively as to normal
        behavior, making persistent anomalies indistinguishable from normal
        operation.
        \item We identify change-point detection as a more suitable use case,
        supported by the observation that forecasting error peaks reliably at
        anomaly boundaries even in the zero-shot regime.
    \end{itemize}

    \subsubsection*{Related Work}

    Classical forecasting-based \gls{mtsad} methods achieve strong performance
    by training directly on the target domain, but at the cost of requiring
    dataset-specific optimization~\cite{sarfraz2024}; in contrast, we study
    whether a pretrained, zero-shot \gls{fm} can offer comparable detection
    capability without any such task-specific training.

    Early work on applying \glspl{fm} to time series anomaly detection focused
    on univariate settings, as surveyed by~\cite{kottapalli2025}.
    The zero-shot transfer of \glspl{fm} to \gls{mtsad} has been explored on
    standard benchmarks such as SMAP and MSL~\cite{olive2026}.
    More recently, ChronosAD~\cite{khan2026} demonstrated that using Google's
    Chronos \gls{fm} as a feature extractor yields promising results across an
    extensive collection of \gls{mtsad} benchmarks, establishing \glspl{fm} as
    a competitive approach when used in a representation-learning role.

    \section{Background}\label{sec:background}
    \subsection{Multivariate Time Series Anomaly Detection}\label{ssec:mtsad}

    We denote by $\mathbf{X} \in \mathbb{R}^{T\times C}$ a multivariate time
    series with $T$
    time steps and $C$ channels (variates).
    The time steps are indexed by $t \in \{ 1, \ldots, T\}$ with a corresponding
    label $y \in \{0,1 \}^T$ indicating benign ($y_t=0$) or anomalous
    behaviour ($y_t=1$).
    The goal of unsupervised \gls{mtsad} is to identify anomalous observations
    $\mathbf{x}_t \in \mathbb{R}^C$ at each time step $t$, based on the
    joint temporal
    behavior across multiple channels.
    This is in contrast to univariate anomaly detection, which only considers
    a single channel in isolation.

    The procedure of unsupervised \gls{mtsad} typically involves two steps:
    first, a model is trained on the multivariate time series data to learn a
    representation of behavior under a benign regime.
    This model can then be leveraged to identify anomalous observations based on
    deviations from the learned representation of normal behavior.
    Classical approaches to \gls{mtsad} include reconstruction-based methods,
    such as autoencoders, and forecasting-based methods~\cite{sarfraz2024}.
    More recently, deep learning-based methods have been proposed, which
    leverage the ability of neural networks to capture complex temporal
    dependencies and cross-channel correlations~\cite{pang2022}.

\begin{figure}[htb]
        \centering
        \includegraphics[width=\textwidth]{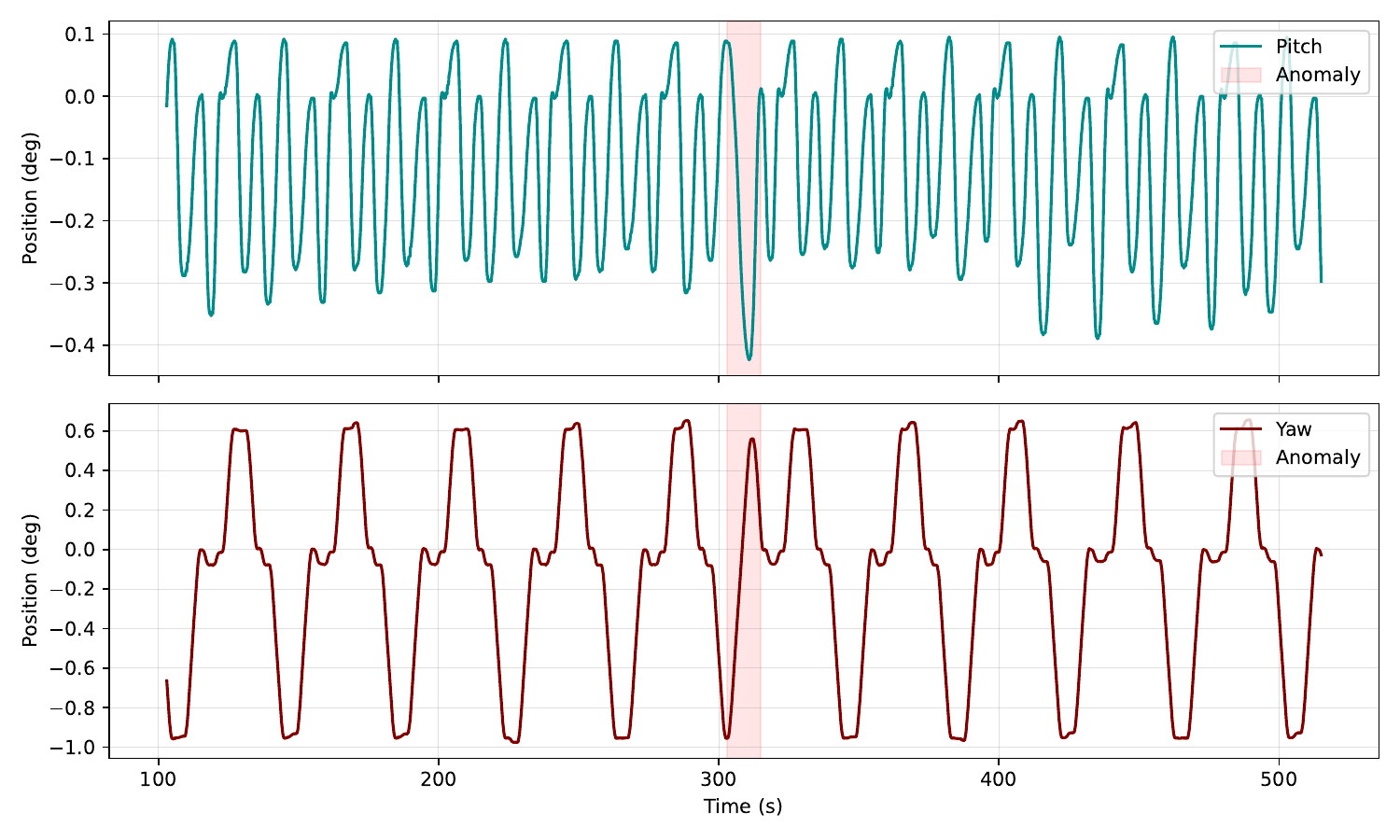}
        \caption{Illustration of anomalies on the \gls{qad} on two selected
        features:
        A multivariate signal is monitored over time and anomalies (highlighted
        in red) are identified
        from deviations of the joint temporal behavior across multiple channels
        rather than from a single variable in isolation.
        Illustration adapted from~\cite{nosrati2026}.}
        \label{fig:mtsad}
    \end{figure}

    To benchmark the performance of \gls{mtsad} methods, several datasets
    from different domains have been
    proposed, including NASA telemetry data~\cite{hundman2018},
    server machine data~\cite{abdulaal2021,su2019}, water treatment plant
    data~\cite{chuadhry2017,goh2017},
    and more recently, the \gls{qad} benchmark~\cite{nosrati2026}, which provides
    a collection of multivariate time series data from a real-world industrial
    system under various operating conditions and fault scenarios.
    An example of a multivariate time series for anomaly detection is
    illustrated in \Cref{fig:mtsad}, showing two selected variates from a trace
    taken from the \gls{qad} benchmark~\cite{nosrati2026}.
    This dataset is not used in this work; however, thanks to its periodic
    behavior, it nicely illustrates the anomalies.

    \subsection{Foundation Models}\label{ssec:foundation-models}
    \glspl{fm} are a class of machine learning models that are trained on
    large-scale
    datasets and can be adapted to a wide range of downstream tasks.
    The most prominent \glspl{fm} are Large Language Models, which have achieved
    remarkable success in natural language processing, but \glspl{fm} have also
    been applied to other modalities, such as image, audio, and video.
    Although the term \gls{fm} is not restricted to a specific
    architecture~\cite{bommasani2021}, the Transformer architecture has become
    the de facto standard for \glspl{fm}~\cite{kottapalli2025}.

    Formally, we define a univariate time series \gls{fm} as a function
    $f_\theta: \mathbb{R}^{w} \to \mathbb{R}^h$ that
    maps an input sequence $x_{t-w:t}$ of context window length $w$ to an output
    sequence $\hat{x}_{t+1 : t+1+h}$ of the horizon length $h$,
    where $\theta$ are the pretrained parameters of the model.

    To make it more generic, we decompose the \gls{fm} $f_\theta$ into
    \[
        f_\theta = \phi_\theta \circ g_\theta
    \]
    an encoder $g_\theta: \mathbb{R}^{w} \to \mathbb{R}^d$ that maps the input
    sequence to a latent representation $v\in \mathbb{R}^d$ and a decoder
    $\phi_\theta: \mathbb{R}^d \to \mathbb{R}^h$ that maps the latent
    representation to the output sequence.

    This work focuses on the application of univariate time series
    \glspl{fm} only~\cite{das2024}, although multivariate time series \glspl{fm}
    are becoming more common, where cross-channel dependencies
    are explicitly modeled or each variate is treated as a separate channel
    in a multivariate time series~\cite{kottapalli2025}.
    However, recent works suggest that treating each channel independently
    suffices for \gls{mtsad} on published
    benchmarks~\cite{garg2022,pinet2026,wenig2024}.

    \section{Methodology}\label{sec:methodology}
    For this exploratory work, we consider two ways of leveraging a univariate
    time series \gls{fm} for multivariate time series anomaly detection,
    which are detailed in the following.
    For both approaches, we use a pretrained univariate time series \gls{fm}
    $f_\theta$ trained to predict the next $h$ time steps of a
    univariate time series given a context window of length $w$.

    \subsection{Forecasting-Based Anomaly Detection}
    \label{sec:forecasting-based-anomaly-detection}
    Similar to classical forecasting-based anomaly detection
    methods~\cite{wu2023,xu2022}, we can leverage a univariate time series
    \gls{fm} to predict future values of each channel in a multivariate time
    series.
    Here we intentionally omit the cross-channel dependencies and treat each
    channel independently, which allows us to use a univariate \gls{fm} for
    multivariate time series anomaly detection.
    Although this approach does not capture the cross-channel dependencies,
    recent works suggest that cross-channel dependencies are not critical for
    anomaly detection on published
    benchmarks~\cite{garg2022,pinet2026,wenig2024}.

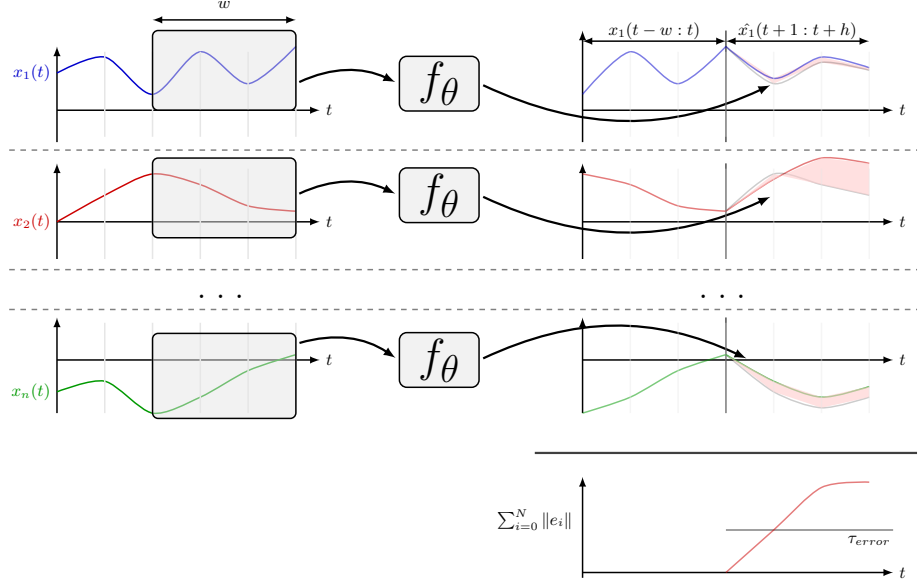
\begin{figure}[t]
        \centering
        \begin{tikzpicture}[x=0.9cm, y=1cm, thick]
    \useasboundingbox (-1.,-5) rectangle (18,6.25); 

    \def\ysep{2.1}
    \def\Tmax{5}
    \def\boxx{\Tmax+3}
    \def\axshift{1} 

    \begin{scope}[shift={(0,2*\ysep)}]
        \draw[->] (0,0) -- (\Tmax+0.5,0);
        \draw[->] (0,-0.6) -- (0,1.2);
        \node at (5.7,0) {$t$};
        \draw[myblue, thick] plot[smooth]
        coordinates{(0,0.7) (1,1.0) (2,0.3) (3,1.1) (4,0.5) (5,1.2)};
        \node[left, myblue] at (0,0.7) {$x_1(t)$};
        \foreach \x in {1,...,\Tmax} \draw[gray!20] (\x,-0.5) -- (\x,{1.1});

        \draw[<->] (2,1.7) -- (5.0,1.7);
        \node at (3.5,2) {$w$};
        \draw[draw, rounded corners=3pt, thick,
            fill=gray!20, fill opacity=0.5] (2,0) rectangle (5,1.5);
        \coordinate (rectRight) at (5, 0.5); 

        \node[draw, rounded corners, fill=boxbg, thick, minimum width=1.5cm, minimum height=0.75cm, align=center]
        (fTheta) at (\boxx,0.5) {\Huge $f_\theta$};
        \draw[->, very thick, shorten >=2pt, shorten <=2pt] (rectRight) to [bend left=25] (fTheta.west);
        \draw[->, very thick, shorten >=2pt, shorten <=2pt] (fTheta.east) to [bend left=-25] (15,0.5);

        \begin{scope}[shift={(\boxx+\axshift,0)}]
            \draw[->] (2,0) -- (8+0.5,0);
            \draw[->] (2,-0.6) -- (2,1.5);
            \node at (8.7,0) {$t$};
            \draw[-,opacity=0.7] (5,-0.6) -- (5,1.5);
            \draw[<->] (2,1.3) -- (5.0,1.3);
            \draw[<->] (5.,1.3) -- (8.0,1.3);
            \node at (3.5,1.5) {$x_1(t-w : t)$};
            \node at (6.5,1.5) {$\hat{x_1}(t+1 : t+h)$};
            \draw[myblue, thick, opacity=0.52] plot[smooth]
            coordinates{(2,0.3) (3,1.1) (4,0.5) (5,1.2)};
            \fill[red!20, opacity=0.6]
            (5,1.2) -- (6,0.6) -- (7,1.0) -- (8,0.8)
            -- (8,0.75) -- (7,0.9) -- (6,0.5) -- (5,1.2)
            -- cycle;

            \draw[myblue, thick, opacity=0.52] plot[smooth] coordinates{
                (5,1.2) (6,0.6) (7,1.0) (8,0.8)};
            \draw[opacity=0.2] plot[smooth] coordinates{
                (5,1.2) (6,0.5) (7,0.9) (8,0.75)};

            \foreach \x in {3,...,8} \draw[gray!20, opacity=0.4] (\x,-0.5) -- (\x,{1.1});
        \end{scope}

    \end{scope}

    \begin{scope}[shift={(0,\ysep)}]
        \draw[->] (0,0) -- (\Tmax+0.5,0);
        \draw[->] (0,-0.7) -- (0,1.1);
        \node at (5.7,0) {$t$};
        \draw[myred, thick] plot[smooth]
        coordinates{(0,0.0) (1,0.5) (2,0.9) (3,0.7) (4,0.3) (5,0.2)};
        \node[left, myred] at (0,0.0) {$x_2(t)$};
        \foreach \x in {1,...,\Tmax} \draw[gray!20] (\x,-0.7) -- (\x,1.0);

        \draw[dashed, opacity=0.5] (-1,1.35) -- (18,1.35);

        \draw[draw, rounded corners=3pt, thick,
            fill=gray!20, fill opacity=0.5] (2,-0.3) rectangle (5,1.2);
        \coordinate (rectRight) at (5, 0.5); 

        \node[draw, rounded corners, fill=boxbg, thick, minimum width=1.5cm, minimum height=0.75cm, align=center]
        (fTheta) at (\boxx,0.5) {\Huge $f_\theta$};
        \draw[->, very thick, shorten >=2pt, shorten <=2pt] (rectRight) to  [bend left=25] (fTheta.west);
        \draw[->, very thick, shorten >=2pt, shorten <=2pt] (fTheta.east) to [bend left=-25] (15,0.5);

        \begin{scope}[shift={(\boxx+\axshift,0)}]
            \draw[->] (2,0) -- (8+0.5,0);
            \draw[->] (2,-0.7) -- (2,1.1);
            \node at (8.7,0) {$t$};
            \draw[-,opacity=0.7] (5,-0.7) -- (5,1.1);
            \draw[myred, thick, opacity=0.52] plot[smooth]
            coordinates{(2,0.9) (3,0.7) (4,0.3) (5,0.2)};
            \draw[myred, thick, opacity=0.52] plot[smooth]
            coordinates{(5,0.2)(6,0.8) (7,1.2) (8,1.1)};
            \draw[opacity=0.2] plot[smooth]
            coordinates{(5,0.2)(6,0.9) (7,0.7) (8,0.5)};

            \fill[red!20, opacity=0.6]
            (5,0.2) -- (6,0.8) -- (7,1.2) -- (8,1.1)
            -- (8,0.5) -- (7,0.7) -- (6,0.9) -- (5,0.2)
            -- cycle;

            \foreach \x in {3,...,8} \draw[gray!20, opacity=0.4] (\x,-0.7) -- (\x,1.0);
        \end{scope}

    \end{scope}

    \begin{scope}[shift={(0,-0.5)}]
        \draw[->] (0,0) -- (\Tmax+0.5,0);
        \draw[->] (0,-1.0) -- (0,0.8);
        \node at (5.7,0) {$t$};
        \draw[mygreen, thick] plot[smooth]
        coordinates{(0,-0.6) (1,-0.4) (2,-1.0) (3,-0.7) (4,-0.2) (5,0.1)};
        \node[left, mygreen] at (0,-0.6) {$x_n(t)$};
        \foreach \x in {1,...,\Tmax} \draw[gray!20] (\x,-1.0) -- (\x,0.7);

        \draw[dashed, opacity=0.5] (-1,0.95) -- (18,0.95);
        \node at (3.5, 1.2) {\Huge \ldots};
        \draw[dashed, opacity=0.5] (-1,1.7) -- (18,1.7);

        \draw[draw, rounded corners=3pt, thick,
            fill=gray!20, fill opacity=0.5] (2,-1.1) rectangle (5,0.5);
        \coordinate (rectRight) at (5, 0.3); 

        \node[draw, rounded corners, fill=boxbg, thick, minimum width=1.5cm, minimum height=0.75cm, align=center]
        (fTheta) at (\boxx,0.0) {\Huge $f_\theta$};
        \draw[->, very thick, shorten >=2pt, shorten <=2pt] (rectRight) to  [bend left=25] (fTheta.west);
        \draw[->, very thick, shorten >=2pt, shorten <=2pt] (fTheta.east) to [bend left=25] (14.5,0.);

        \begin{scope}[shift={(\boxx+\axshift,0)}]
            \draw[->] (2,0) -- (8+0.5,0);
            \draw[->] (2,-1.0) -- (2,0.8);
            \node at (8.7,0) {$t$};
            \draw[-,opacity=0.7] (5,-1.0) -- (5,0.8);
            \draw[mygreen, thick, opacity=0.52] plot[smooth]
            coordinates{ (2,-1.0) (3,-0.7) (4,-0.2) (5,0.1)  };
            \draw[opacity=0.2] plot[smooth]
            coordinates{ (5,0.1) (6,-0.6) (7,-0.9) (8,-0.7)};
            \draw[mygreen, thick, opacity=0.7] plot[smooth]
            coordinates{ (5,0.1) (6,-0.4) (7,-0.7) (8,-0.5)};

            \fill[red!20, opacity=0.6]
            (5,0.1) -- (6,-0.6) -- (7,-0.9) -- (8,-0.7)
            -- (8,-0.5) -- (7,-0.7) -- (6,-0.4) -- (5,0.1)
            -- cycle;

            \foreach \x in {3,...,8} \draw[gray!20, opacity=0.4] (\x,-1.0) -- (\x,0.7);

        \end{scope}

        \node at (14, 1.2) {\Huge \ldots};

    \end{scope}

    \begin{scope}[shift={(0,-4.5)}]
        \begin{scope}[shift={(\boxx+\axshift,0)}]
            \draw[->] (2,0) -- (8+0.5,0);
            \draw[->] (2,-0.1) -- (2,1.8);
            \draw[opacity=0.75, very thick] (1,2.25) -- (9,2.25);
            \draw[opacity=0.5] (5,0.8) -- (8.5,0.8);
            \node at (8.0,0.6) {$\tau_{{error}}$};
            \node at (8.7,0) {$t$};
            \node at (1,1) {$\sum_{i=0}^{N} \|e_i\|$};
            \draw[myred, thick, opacity=0.52] plot[smooth]
            coordinates{(5,0)(6,0.8) (7,1.6) (8,1.7)};
        \end{scope}
    \end{scope}
\end{tikzpicture}
        \caption{Forecasting-based anomaly detection with a \gls{fm}. For each
        channel, a context window of length $w$ is mapped by $f_\theta$ to a
        forecast horizon of length $h$, and anomalies are assessed from the
        discrepancy between predicted and observed future values. The lower
        panel illustrates the aggregation of per-channel prediction errors into
        a joint score that can be compared against a threshold $\tau_{error}$.}
        \label{fig:forecasting-based-anomaly-detection}
    \end{figure}

    As illustrated in \Cref{fig:forecasting-based-anomaly-detection}, we detect
    anomalies by comparing the predicted future values $\hat{X}_t$ with the
    observed future values $X_t$. For each time step $t$, we compute a joint
    prediction error $e_t$ by aggregating the per-channel errors
    $e_{c,t} = \| \hat{X}_{c,t} - X_{c,t} \|$ over all channels,
    \[ e_t = \sum^{C}_{c=1} e_{c,t}. \]
    To identify anomalous observations, the prediction error is then
    thresholded against a predefined value $\tau_{error}$
    \[
        \hat{y}_t = \mathbb{I}(e_t > \tau_{error}),
    \]
    where $\hat{y}_t$ is the predicted label for time step $t$ and $\mathbb{I}$
    is the indicator function.

    \subsection{Embedding-Based Anomaly Detection}
    \label{sec:embedding-based-anomaly-detection}

\begin{figure}
        \centering
        \begin{tikzpicture}[x=0.9cm, y=1cm, thick]
    \useasboundingbox (-1.,-2.75) rectangle (17,6.25); 

    \def\ysep{2.1}
    \def\Tmax{5}
    \def\boxx{\Tmax+3}
    \def\axshift{1} 

    \begin{scope}[shift={(0,2*\ysep)}]
        \draw[->] (0,0) -- (\Tmax+0.5,0);
        \draw[->] (0,-0.6) -- (0,1.2);
        \node at (5.7,0) {$t$};
        \draw[myblue, thick] plot[smooth]
        coordinates{(0,0.7) (1,1.0) (2,0.3) (3,1.1) (4,0.5) (5,1.2)};
        \node[left, myblue] at (0,0.7) {$x_1(t)$};
        \foreach \x in {1,...,\Tmax} \draw[gray!20] (\x,-0.5) -- (\x,{1.1});

        \draw[<->] (2,1.7) -- (5.0,1.7);
        \node at (3.5,2) {$w$};
        \draw[draw, rounded corners=3pt, thick,
            fill=gray!20, fill opacity=0.5] (2,0) rectangle (5,1.5);
        \coordinate (rectRight) at (5, 0.5); 

        \node[draw, rounded corners, fill=boxbg, thick, minimum width=1.5cm, minimum height=0.75cm, align=center]
        (fTheta) at (\boxx,0.5) {\Huge $g_\theta$};
        \draw[->, very thick, shorten >=2pt, shorten <=2pt] (rectRight) to [bend left=25] (fTheta.west);
        \draw[->, very thick, shorten >=2pt, shorten <=2pt] (fTheta.east) to (11,0.5);

        \begin{scope}[shift={(11,0)}]
            \draw[thick] (0,0) rectangle (1,1);
            \draw[thick] (1,0) rectangle (2,1);
            \draw[thick] (2,0) rectangle (3,1);
            \draw[thick] (3,0) rectangle (4,1);
            \node at (0.5,0.5) {$v_{00}$};
            \node at (1.5,0.5) {$v_{01}$};
            \node at (2.5,0.5) {$\ldots$};
            \node at (3.5,0.5) {$v_{0d}$};
            \node at (5,0.5) {$z_{t,0}$};
        \end{scope}
    \end{scope}

    \begin{scope}[shift={(0,\ysep)}]
        \draw[->] (0,0) -- (\Tmax+0.5,0);
        \draw[->] (0,-0.7) -- (0,1.1);
        \node at (5.7,0) {$t$};
        \draw[myred, thick] plot[smooth]
        coordinates{(0,0.0) (1,0.5) (2,0.9) (3,0.7) (4,0.3) (5,0.2)};
        \node[left, myred] at (0,0.0) {$x_2(t)$};
        \foreach \x in {1,...,\Tmax} \draw[gray!20] (\x,-0.7) -- (\x,1.0);
        \draw[dashed, opacity=0.5] (-1,1.35) -- (17,1.35);

        \draw[draw, rounded corners=3pt, thick,
            fill=gray!20, fill opacity=0.5] (2,-0.3) rectangle (5,1.2);
        \coordinate (rectRight) at (5, 0.5); 
        \node[draw, rounded corners, fill=boxbg, thick, minimum width=1.5cm, minimum height=0.75cm, align=center]
        (fTheta) at (\boxx,0.5) {\Huge $g_\theta$};

        \draw[->, very thick, shorten >=2pt, shorten <=2pt] (rectRight) to  [bend left=25] (fTheta.west);
        \draw[->, very thick, shorten >=2pt, shorten <=2pt] (fTheta.east) to (11,0.5);

        \begin{scope}[shift={(11,0)}]
            \draw[thick] (0,0) rectangle (1,1);
            \draw[thick] (1,0) rectangle (2,1);
            \draw[thick] (2,0) rectangle (3,1);
            \draw[thick] (3,0) rectangle (4,1);
            \node at (0.5,0.5) {$v_{10}$};
            \node at (1.5,0.5) {$v_{11}$};
            \node at (2.5,0.5) {$\ldots$};
            \node at (3.5,0.5) {$v_{1d}$};
            \node at (5,0.5) {$z_{t,1}$};
        \end{scope}
    \end{scope}

    \begin{scope}[shift={(0,-0.5)}]
        \draw[->] (0,0) -- (\Tmax+0.5,0);
        \draw[->] (0,-1.0) -- (0,0.8);
        \node at (5.7,0) {$t$};
        \draw[mygreen, thick] plot[smooth]
        coordinates{(0,-0.6) (1,-0.4) (2,-1.0) (3,-0.7) (4,-0.2) (5,0.1)};
        \node[left, mygreen] at (0,-0.6) {$x_n(t)$};
        \foreach \x in {1,...,\Tmax} \draw[gray!20] (\x,-1.0) -- (\x,0.7);

        \draw[draw, rounded corners=3pt, thick,
            fill=gray!20, fill opacity=0.5] (2,-1.1) rectangle (5,0.5);
        \coordinate (rectRight) at (5, 0.3); 

        \node[draw, rounded corners, fill=boxbg, thick, minimum width=1.5cm, minimum height=0.75cm, align=center]
        (fTheta) at (\boxx,0.0) {\Huge $g_\theta$};
        \draw[->, very thick, shorten >=2pt, shorten <=2pt] (rectRight) to [bend left=25] (fTheta.west);

        \begin{scope}[shift={(11,-0.5)}]
            \draw[thick] (0,0) rectangle (1,1);
            \draw[thick] (1,0) rectangle (2,1);
            \draw[thick] (2,0) rectangle (3,1);
            \draw[thick] (3,0) rectangle (4,1);
            \node at (0.5,0.5) {$v_{n0}$};
            \node at (1.5,0.5) {$v_{n1}$};
            \node at (2.5,0.5) {$\ldots$};
            \node at (3.5,0.5) {$v_{nd}$};
            \node at (5,0.5) {$z_{t,n}$};
        \end{scope}

        \draw[->, very thick, shorten >=2pt, shorten <=2pt] (fTheta.east) to (11,0.);

        \draw[dashed, opacity=0.5] (-1,0.95) -- (17,0.95);
        \node at (3.5, 1.2) {\Huge \ldots};
        \node at (13, 1.2) {\Huge \ldots};
        \draw[dashed, opacity=0.5] (-1,1.7) -- (17,1.7);
    \end{scope}

    \node at (13, -2.5) {\large $z_t = [z_{t,0}, z_{t,1}, \ldots, z_{t,n}] \in \mathbb{R}^d$};

\end{tikzpicture}
        \caption{Embedding-based anomaly detection with a \gls{fm}.
        Each univariate input segment is processed by the encoder $g_\theta$ to
        produce channel-wise latent representations, which are then combined
        into a joint embedding vector $z_t \in \mathbb{R}^{C \cdot d}$ for the
        multivariate observation at time $t$.
        This representation can subsequently be used by downstream anomaly
        scoring methods to detect abnormal system states in a
        zero-shot setting.}
        \label{fig:emb-based-anomaly-detection}
    \end{figure}
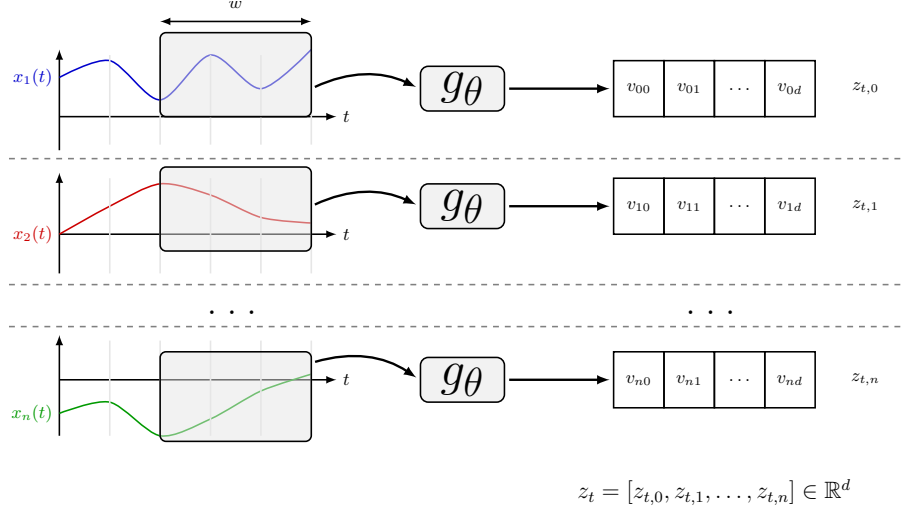

    In contrast to the forecasting-based approach, we can also leverage the
    latent representations of a univariate time series \gls{fm} for anomaly
    detection.
    Here, we use the encoder $g_\theta$ of the \gls{fm} to map each channel of a
    multivariate time series to a latent
    representation $v_{c,t} \in \mathbb{R}^d$
    for each channel $c$ and time step $t$.
    These per-channel representations serve as input features for a downstream
    anomaly detection method.
    The latent representations of all channels are then concatenated to form a
    joint embedding vector $z_t \in \mathbb{R}^{C \cdot d}$ for the multivariate
    observation at time step $t$.
    This joint-embedding vector can then be used as input to a downstream
    anomaly detection method, such as an \gls{if}~\cite{liu2008}, to identify
    anomalous observations based on their distance to the learned representation
    of normal behavior.
    The overall procedure of extracting the joint embedding vector is
    illustrated in \Cref{fig:emb-based-anomaly-detection}.

    Following the standard procedure of unsupervised anomaly detection methods,
    we can then use the joint-embedding vectors $z_t$ to train an \gls{if} on
    the normal operating regime of the system and use it to identify anomalous
    observations based on their distance to the learned representation of normal
    behavior.

    \section{Experiments \& Results}\label{sec:experiments-results}

    \paragraph{Experimental Setup.}
    For all experiments, we use TimesFM~\cite{das2024} (version 1.2.0) via its
    publicly available implementation\footnote{%
        \url{https://github.com/google-research/timesfm}
    } as the univariate time series \gls{fm}. TimesFM is a Transformer-based
    model pretrained on a large corpus of univariate time series data from
    various domains.
    Although it is possible to fine-tune the model on the target domain, we
    explicitly focus on the zero-shot regime in this work, where the pretrained
    model is directly applied to the target domain without any further training
    or adaptation.

    We use a context length of $w=100$, consistent with related
    work~\cite{uray2026},
    and a forecast horizon of $h=32$, chosen as a trade-off between
    TimesFM's long-horizon forecast capability and per-step prediction accuracy.
    All experiments are conducted on a single NVIDIA GeForce RTX 3090 with 24 GB
    of VRAM and an Intel(R) Core(TM) i9-10980XE CPU @ 3.00GHz.

    \paragraph{Datasets.}
    We evaluate the proposed approaches on the \gls{swat}
    benchmark~\cite{goh2017},
    which is a publicly available dataset that contains multivariate time series
    data from a real-world water treatment plant under various operating
    conditions and fault scenarios.
    The dataset contains 51 channels and a total of 11 days of data, with 7 days
    of normal operation and 4 days of anomalous operation.
    The dataset is split into a training set of 7 days and a test set of 4 days,
    with the anomalous operation occurring in the test set.
    The dataset is preprocessed by normalizing each channel to have zero mean
    and unit variance, and by removing any missing values.

    \paragraph{Metrics.}
    We evaluate the performance of the proposed approaches using the F1 score,
    which is a commonly used metric for anomaly detection that balances
    precision and recall.
    The threshold $\tau_{error}$ for the prediction error in the
    forecasting-based approach is
    optimized on the test set to maximize the F1 score.

    \subsection{Qualitative View on Forecasting-Based Anomaly Detection}
    \label{ssec:forecasting-based-anomaly-detection}
\begin{figure}[t]
        \centering
        \begin{subfigure}[b]{1\textwidth}
            \centering
            \includegraphics[width=\linewidth]{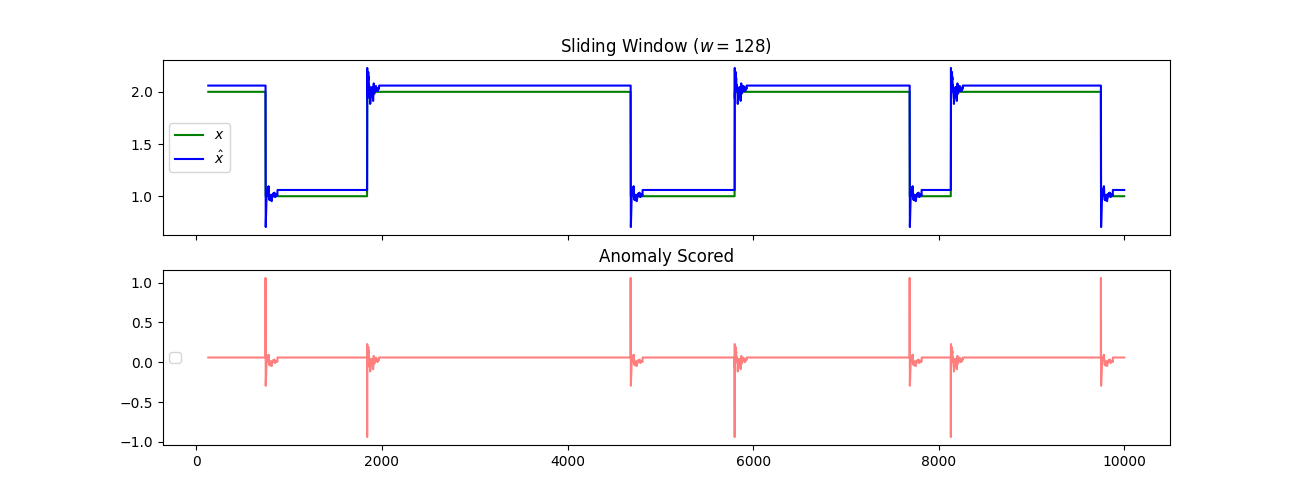}
            \caption{Exemplary forecast on a representative channel with its
            per-channel error profile on benign data; sharp transitions in the
            signal cause transient error spikes even in the normal regime.}
            \label{fig:score-univariate}
        \end{subfigure}
        \hfill
        \begin{subfigure}[b]{1\textwidth}
            \centering
            \includegraphics[width=\linewidth]{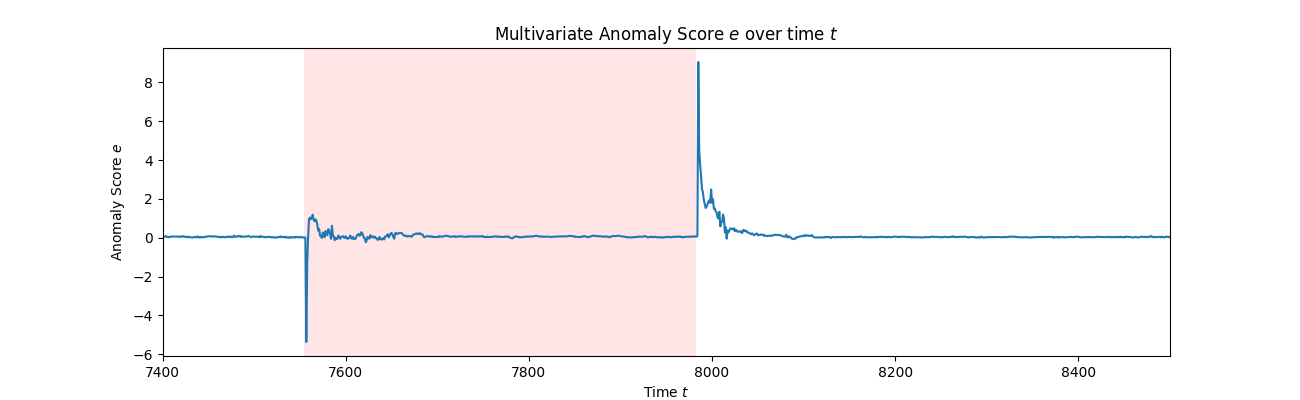}
            \caption{Aggregated multivariate score profile over a window
            containing an anomalous segment; the score spikes at the entry and
            exit boundaries but collapses to near zero throughout the anomaly
            itself.}
            \label{fig:score-multivariate}
        \end{subfigure}
        \caption{Anomaly scoring on the \gls{swat} dataset using per-channel
        forecasts from a \gls{fm}, with per-channel errors summed into
        a joint score.}
        \label{fig:score_forecasting}
    \end{figure}

    \Cref{fig:score_forecasting} illustrates the per-channel and aggregated
    prediction error on a representative excerpt of the \gls{swat} dataset.

    As shown in \Cref{fig:score-univariate}, the \gls{fm} reconstructs the
    benign signal accurately, keeping the prediction error consistently low.
    The quantitative results are correspondingly poor, with an F1 score close
    to zero, confirming that the forecasting-based approach fails to detect
    persistent anomalies.

    The multivariate score profile in \Cref{fig:score-multivariate} reveals
    the underlying cause.
    The prediction error is low during benign operation, spikes sharply at the
    transitions into and out of the anomalous region, but returns to near zero
    throughout the anomaly itself.
    This spike is transient: it persists only while the sliding window of
    length $w$ straddles the boundary; once the window lies entirely within
    the anomalous segment, the \gls{fm} adapts to the new dynamics and the
    score collapses.
    The model is therefore too effective at capturing temporal structure —
    it tracks anomalous behavior just as faithfully as normal behavior,
    rendering persistent anomalies indistinguishable from normal operation.
    Error peaks at anomaly boundaries do, however, make the \gls{fm} a
    natural candidate for change-point detection.

    \subsection{Embedding-Based Anomaly Detection}%
    \label{ssec:embedding-based-anomaly-detection}

    We apply the encoder $g_\theta$ independently to each variate and, for
    each channel, concatenate the intermediate representations from all
    transformer blocks into a single feature vector.
    These per-channel vectors are then concatenated across channels to form
    the joint embedding $z_t$ per time step, as described in
    \Cref{sec:embedding-based-anomaly-detection}.
    An \gls{if} is trained on embeddings from the normal operating regime and
    used to score the test set.

\begin{figure}[t]
        \centering
        \includegraphics[height=0.4\textheight]{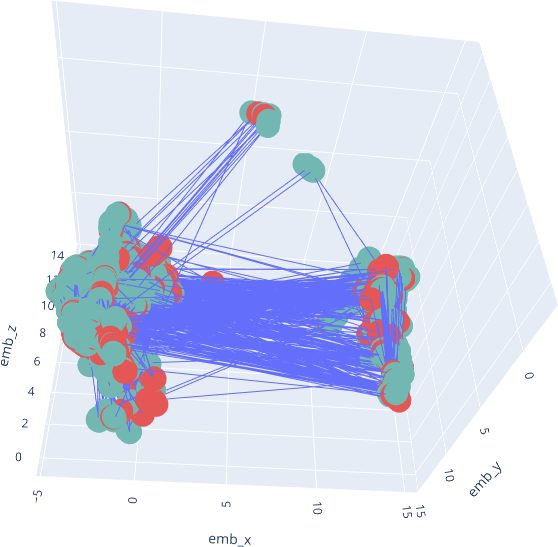}
        \caption{PCA projection of TimesFM embeddings on the \gls{wadi}
        benchmark, colored by operating regime (normal vs.\ anomalous).
        The partial cluster separation indicates that zero-shot representations
        carry some regime-discriminative structure, but the overlap explains
        why the downstream \gls{if} fails to reliably separate the two classes.}
        \label{fig:emb_wadi}
    \end{figure}

    \Cref{fig:emb_wadi} shows the embeddings extracted from the
    \gls{wadi} benchmark~\cite{chuadhry2017}, projected to three dimensions
    via PCA\@.
    Normal and anomalous embeddings form partially overlapping clusters,
    suggesting that the \gls{fm} encodes some structure correlated with the
    operating regime, though the separation is not pronounced enough for
    reliable detection in the full embedding space.

    \Cref{tab:results_embedding} reports F1 scores on the \gls{swat} benchmark.
    The proposed TimesFM + \gls{if} pipeline achieves an F1 of 0.480, well
    below all baselines — including simple heuristics such as L2-norm (0.782)
    and the best-performing method, PCA Error (0.833).

    \begin{table}
        \centering
        \caption{F1 scores on the \gls{swat} benchmark.
        Baselines are taken from~\cite{sarfraz2024} and span simple heuristics
            (top group), lightweight neural methods (middle group), and the
            proposed
            zero-shot TimesFM + \gls{if} pipeline (bottom).}

        \vspace*{1em}
        \label{tab:results_embedding}
        {\setlength{\tabcolsep}{6pt}
            \begin{tabular*}{0.6\textwidth}{@{\extracolsep{\fill}}lc@{}}
            \toprule
            \textbf{Method}          & \textbf{SWaT} \\
            \midrule
            Random                   & 0.218 \\
            Sensor range deviation   & 0.231 \\
            L2-norm                  & 0.782 \\
            1-NN distance            & 0.782 \\
            PCA Error                & \textbf{0.833} \\
            \midrule
            1-Layer MLP              & 0.771 \\
            Single block MLPMixer    & 0.780 \\
            Single Transformer block & 0.787 \\
            1-Layer GCN-LSTM         & \underline{0.829} \\
            \midrule
            TimesFM + \gls{if}       & 0.480 \\
            \bottomrule
            \end{tabular*}}
    \end{table}

    The large gap to even the simplest baselines indicates that zero-shot
    embeddings from a univariate forecasting \gls{fm} do not provide a
    sufficiently discriminative representation for downstream anomaly detection
    on industrial data.
    This is consistent with the qualitative observation in
    \Cref{fig:emb_wadi}: although some cluster structure is visible in the
    projected space, the embeddings of normal and anomalous segments overlap
    substantially, leaving the \gls{if} unable to separate them reliably.

    \section{Discussion \& Future Work}\label{sec:discussion}

    Both strategies fail for the same underlying reason, but it is not the one
    intuitively expected: the \gls{fm} does not lack domain knowledge — it
    generalizes too well.
    In the forecasting-based approach, the model adapts to anomalous dynamics
    within a single sliding window, keeping the prediction error low even deep
    inside an anomalous segment.
    In the embedding-based approach, the same adaptability causes normal and
    anomalous segments to produce overlapping representations.
    The core issue is therefore not a failure to generalize, but a failure to
    \emph{specialize}: a model pre-trained for accurate forecasting optimizes
    for reconstruction fidelity rather than anomaly sensitivity, and those two
    objectives are fundamentally at odds for persistent anomalies.

    This distinction is reflected in the baseline comparison: PCA Error, the
    strongest baseline in \Cref{tab:results_embedding}, fits a single global
    subspace of normal behavior over the entire training set, whereas the
    \gls{fm} re-adapts locally within each sliding window.
    A persistent anomaly therefore remains a persistent outlier with respect to
    a fixed global model, while it is gradually absorbed into the \gls{fm}'s
    local forecast.

    This negative result stands in contrast to ChronosAD~\cite{khan2026}, which
    reports strong performance using embeddings from a different \gls{fm}
    (Chronos) as a feature extractor across a broad collection of \gls{mtsad}
    benchmarks. We see several possible explanations for this discrepancy,
    which we cannot yet disambiguate from our results alone. First, Chronos and
    TimesFM differ in architecture and pretraining corpus, and may therefore
    encode different inductive biases with respect to anomaly-relevant
    structure. Second, ChronosAD evaluates across a wide range of benchmarks,
    whereas our embedding analysis is limited to \gls{swat} only; it is possible
    that our chosen datasets are simply less favorable to this
    representation-based approach. Third, downstream detector choice may
    matter: we pair embeddings with a single Isolation Forest, while
    ChronosAD's pipeline may use a different or better-tuned scoring method.
    Disentangling these factors --- \gls{fm} choice, benchmark choice, and
    detector choice --- is an important direction for future work.

    This boundary-sensitivity property points to a more appropriate use case.
    Because the forecasting error spikes sharply at the transitions into and
    out of anomalous segments (\Cref{fig:score_forecasting}), the \gls{fm}
    reliably signals distributional changes even in the zero-shot regime.
    This makes it a natural candidate for \emph{change-point detection}, a
    related but distinct task from anomaly detection, where the goal is to
    locate the onset of a regime shift rather than to label every anomalous
    time step.

    \paragraph{Future Work.}
    Fine-tuning the \gls{fm} on target-domain data, or even a few-shot
    adaptation, could shift the model toward anomaly sensitivity by exposing it
    to domain-specific normal behavior.
    A complementary direction is to replace the univariate \gls{fm} with a
    genuinely multivariate one that explicitly models cross-channel
    dependencies.
    Although current benchmarks show limited benefit from cross-channel modeling
    in MTSAD~\cite{pinet2026}, real-world industrial systems often exhibit
    strong inter-sensor correlations that a multivariate \gls{fm} could exploit.
    Finally, the boundary-detection property observed here suggests that
    \glspl{fm} may support root cause analysis~\cite{mueller2025}: by
    identifying which channels exhibit the sharpest error spike at a transition,
    one could localize the origin of a fault without any task-specific training.

    \paragraph{Limitations.}
    The main limitation of this work is its narrow experimental scope: we
    evaluate
    a single univariate \gls{fm} on a single benchmark dataset, which constrains
    the
    generalizability of our conclusions. We further restrict our study to the
    zero-shot regime; performance may differ substantially under few-shot or
    fine-tuning settings. Likewise, we pair the \gls{fm} with a single class of
    downstream anomaly detectors, leaving open whether other detection methods
    would
    yield different results. Finally, our setup assumes regularly sampled time
    series and does not address the sparse, irregularly sampled data frequently
    encountered in industrial practice~\cite{uray2026}.

    \section{Conclusion}\label{sec:conclusion}
    In this work, we investigated whether univariate time series \glspl{fm} can
    be
    applied to \gls{mtsad} in a zero-shot regime, without any task-specific
    training. Using TimesFM on the \gls{swat} benchmark, we evaluated two
    complementary
    strategies: employing the \gls{fm} as a per-feature forecaster with
    thresholded
    prediction errors, and as an embedder whose intermediate representations
    feed
    standard outlier detectors. Our results show that neither strategy is
    competitive with established baselines. The qualitative analysis attributes
    this
    to the \gls{fm} being too effective at capturing temporal dynamics: it
    produces
    low forecasting error even within fully anomalous windows, rendering
    persistent
    anomalies indistinguishable from normal behavior.

    At the same time, this very property suggests a more promising use case.
    Since
    the forecasting error peaks at the transitions into and out of anomalous
    segments, \glspl{fm} reliably respond to changes in the underlying data
    distribution, making them strong candidates for change-point detection
    rather
    than for detecting persistent anomalies.

    Future work should explore fine-tuning the \gls{fm} on target-domain data to
    overcome the limitations of the zero-shot regime, evaluate genuinely
    multivariate \glspl{fm} that capture cross-channel dependencies, and
    extend the analysis to a broader range of benchmarks and downstream
    detection
    methods. Investigating the use of \glspl{fm} for change-point detection and
    root-cause analysis constitutes a further promising direction.

    \begin{credits}
        \subsubsection{\ackname}
        The financial support by the Austrian Federal Ministry of Economy,
        Energy and Tourism, the National Foundation for Research, Technology and
        Development and the Christian Doppler Research Association is gratefully
        acknowledged.

        \subsubsection{\discintname}
        The authors have no competing interests to declare that are relevant
        to the content of this article.
    \end{credits}

    \bibliographystyle{splncs04}
    \bibliography{2026-eurocast}
\end{document}